# Search for Choquet-optimal paths under uncertainty

**Lucie Galand**
LIP6-UPMC
104 Av du Président Kennedy
75016 Paris, France

**Patrice Perny**
LIP6-UPMC
104 Av du Président Kennedy
75016 Paris, France

## Abstract

Choquet expected utility (CEU) is one of the most sophisticated decision criteria used in decision theory under uncertainty. It provides a generalisation of expected utility enhancing both descriptive and prescriptive possibilities. In this paper, we investigate the use of CEU for path-planning under uncertainty with a special focus on robust solutions. We first recall the main features of the CEU model and introduce some examples showing its descriptive potential. Then we focus on the search for Choquet-optimal paths in multivalued implicit graphs where costs depend on different scenarios. After discussing complexity issues, we propose two different heuristic search algorithms to solve the problem. Finally, numerical experiments are reported, showing the practical efficiency of the proposed algorithms.

## 1 INTRODUCTION

An important source of complexity in practical applications of problem solving methods developed in AI is the imperfect knowledge of the real problem to deal with. This is particularly true in path-planning problems where the map is not always the territory. When seeking the optimal path from a given node to a goal node, several uncertainty factors might indeed increase the complexity of the optimization task. Firstly, the consequences of the actions might not be certain, which can be modeled by non-deterministic transitions between states. Secondly, the current state might not be known exactly (partial observability), which requires maintaining beliefs on possible states revised during the search. These problems are widely discussed in the litterature on MDPs and POMDPs, see e.g. Puterman (1994), Kaebling et al. (1999).

Beside these sources of complexity, the cost transitions between states might also be uncertain. This eventuality has motivated work aimed at revisiting, under uncertainty, the shortest path problem in a state space graph, and its classical resolution with the $A^*$ algorithm. For example, Wellman and Wurman consider a case where the costs are time dependent and representable by random variables (Wellman et al., 1995). They introduce the $SDA^*$ algorithm to determine the preferred paths according to the stochastic dominance partial order. Moreover, an extension of this algorithm specifically designed to cope with both uncertainty and multiple criteria is proposed in (Wurman and Wellman, 1996).

Another way of introducing uncertainty in costs is to consider a set of plausible scenarios, each bringing a different valuation to transitions and therefore to solution paths. This is the natural formulation when the costs depend on exogenous variables not controlled by the decision maker, these variables having an overall impact on the graph (e.g. transfer times in a city depending on the weather, security of moves depending on enemy positions, asset values depending on market evolution). The introduction of scenarios implicitly defines a family of possible instances of the same problem, all sharing the same feasible solutions, but with different views on the possible costs. In such problems, the aim is to seek for "robust" solutions, i.e. solutions yielding a "reasonable" cost in all plausible instances of the problem. This robustness idea has actively developed in discrete optimization since the publication of the book by Kouvelis and Yu (1997), which considers several criteria imported from decision theory under total uncertainty (e.g. min-max, min-max regret) to define the absolute or relative robustness of a solution. Under these criteria, the shortest path problem becomes NP-hard, as do many other polynomially solvable problems (see e.g., Aron and van Hentenryck (2002) for robust spanning tree problems), thus bringing new algorithmic challenges. In the same vein, alternative models such as Lorenz dominance and ordered weighted averages have been introduced and justified to model robustness in (Perny and Spanjaard, 2003), as well as a multiobjective search algorithm to determine robust solution paths in state space graphs under total uncertainty.

One major limitation of these robustness criteria is overpessimism (worst case analysis) in cost aggregation, making them not sufficiently discriminating. When information about the relative plausibility of scenarios is available, robustness criteria can be refined using models introduced in decision theory under uncertainty and risk to convey pessimism, prudence, risk-aversion or uncertainty aversion in the comparisons of acts. This idea is exploited in (Perny et al., 2007) for path-planning under risk (when probabilities of scenarios are known), where a multiobjective heuristic search algorithm is proposed for the determination of optimal solution paths with respect to second-order stochastic dominance, or expected utility for risk-averse agents. However, in some situations, these models do not apply directly, either because the objective probabilities of scenarios are not known, or because observed preferences do not match EU theory with respect to any probabilities. An example of such a problematic situation, inspired by the so-called Ellsberg's urn example (Ellsberg, 1961), is presented below in the context of path planning:

**Example 1** *Consider a problem with 3 scenarios ($S = \{1, 2, 3\}$) and assume that objective probabilities of scenarios are imprecisely known and defined by $p_1 = 1/3$ and $p_2 + p_3 = 2/3$. Consider now 4 solution paths $P_1, P_2, P_3, P_4$ to reach a goal node from the initial node with the following costs $c(P_i, s_j)$:*

|       | 1   | 2   | 3   |
|-------|-----|-----|-----|
| $P_1$ | 0   | 100 | 100 |
| $P_2$ | 100 | 0   | 100 |
| $P_3$ | 0   | 100 | 0   |
| $P_4$ | 100 | 0   | 0   |

*On the one hand, a decision maker who is averse to uncertainty might prefer $P_1$ to $P_2$ because $P_1$ has probability $1/3$ of reaching the goal for free, whereas $P_2$ might have a cost of 100 for sure. On the other hand, the same decision maker might prefer $P_4$ to $P_3$ because $P_4$ has probability $2/3$ of reaching the goal for free, whereas $P_3$ might have only probability $1/3$. Although such preferences are natural they cannot be described by the EU model. Indeed $P_1 \succ P_2$ implies $p_2 + p_3 > p_1 + p_3$ whereas $P_4 \succ P_3$ implies $p_1 > p_2$. Note that these inequalities are contradictory. This makes it impossible to assume that subjective probabilities are implicitly assigned to states by the decision maker. Trying to reveal them would be meaningless.*

The impossibility of revealing probabilities through the EU model in some situations has led to the introduction of alternative numerical representations of beliefs in events by a capacity function, a less constraining representation of uncertainty that relaxes the additivity assumption. The standard decision model associated to capacities is based on the use of the Choquet integral, following Schmeidler (1989), who gives the foundations of the Choquet Expected Utility criterion (CEU). Despite its descriptive appeal, until now, CEU has not been considered in the field of path-planning under uncertainty. The aim of this paper is to fill this gap and to complete the previous studies by investigating the potential of CEU in the search of robust paths. The paper is organized as follows. In Section 2 we recall the main features of the CEU model and introduce some examples showing its descriptive potential in the context of search under uncertainty; then we discuss complexity issues concerning the search of Choquet-optimal paths. In Section 3 we propose two different heuristic search algorithms to solve the problem. Finally, numerical experiments are reported in Section 4, showing the practical efficiency of the proposed algorithms.

## 2 PROBLEM FORMULATION

### 2.1 Notations and Definitions

We consider a state space graph $G = (N, A)$ where $N$ is a finite set of nodes (possible states), and $A$ is a set of arcs representing feasible transitions between nodes. Formally, we have $A = \{(n, n') : n \in N, n' \in S(n)\}$ where $S(n) \subseteq N$ is the set of all successors of node $n$ (nodes that can be reached from $n$ by a feasible elementary transition). We denote $\mathcal{P}(n, n')$ the set of all paths linking $n$ to $n'$, and $\mathcal{P}(n, N')$ the set of all paths from $n$ to any node $n' \subseteq N'$ (with $N' \subseteq N$). We call *solution path* a path from $s$ to a goal node $\gamma \in \Gamma$ (i.e. a path in $\mathcal{P}(s, \Gamma)$). Throughout the paper, we assume that there exists at least one finite length solution path. We consider a finite set $S = \{1, \ldots, m\}$ of possible scenarios, each having possibly a different impact on the transition costs, and a scenario-dependent valuation $c : A \times S \to \mathbb{R}_+$ where $c(a, s)$ is the cost of the transition represented by $a$ in scenario $s$. By abuse of notation, $c(a)$ denotes the cost vector $(c(a, 1), \ldots, c(a, m))$. The costs over a path are assumed to be additive, which allows valuation $c$ to be extended from arcs to paths by setting, for any path $P$ and any scenario $s$, $c(P, s) = \sum_{a \in P} c(a, s)$. A cost vector $x = (x_1, x_2, \ldots, x_m) \in \mathbb{R}_+^m$ is associated to each path $P$ in $G$ such that $x_i = c(P, i)$, hence, the comparison of paths reduces to the comparison of their associated cost vectors. In the sequel we assume that, for all scenarios, the cost of each solution path is bounded by a positive constant $M$. Throughout the paper, we will consider a weak preference over paths (or cost vectors), denoted $\succsim$. For any pair of paths $P$ and $P'$ with respective costs $x$ and $x'$ in $\mathbb{R}_+^m$ we will use notation $P \succsim P'$ or $x \succsim x'$ to say $P$ is at least as good as $P'$. As usual relation $P \sim P'$ (or $x \sim x'$) represents indifference between the two paths and corresponds to $P \succsim P'$ and $P' \succsim P$. Finally, relation $P \succ P'$ (or $x \succ x'$) represents strict preference and corresponds to $P \succsim P'$ and $not(P' \succsim P)$.

## 2.2 Capacities and the Choquet Integral

The Choquet integral (Choquet, 1953) is used in decision theory to generalize the notion of expectation when beliefs in events are represented by a *capacity*, i.e. a set function $v : 2^S \to [0,1]$ such that $v(\emptyset) = 0$, $v(S) = 1$ and $\forall A, B \in 2^S$ such that $A \subseteq B$, $v(A) \leq v(B)$. For any event $A \subseteq S$, $v(A)$ represents the plausibility of event $A$. The capacity $v$ is said to be *convex* (or supermodular) when $v(A \cup B) + v(A \cap B) \geq v(A) + v(B)$ for all $A, B \subseteq S$, and it is said to be *concave* (or submodular) when $v(A \cup B) + v(A \cap B) \leq v(A) + v(B)$ for all $A, B \subseteq S$. To any capacity $v$, we can associate a dual capacity $\bar{v}$ defined by $\bar{v}(A) = 1 - v(S \setminus A)$ for all $A \subseteq S$. It is well known that $\bar{v}$ is concave if and only if $v$ is convex and vice-versa. When $v$ is concave, we have $v(A) + v(S \setminus A) \geq 1$, hence $\bar{v}(A) \leq v(A)$. In this case the core of $\bar{v}$ defined by $core(\bar{v}) = \{P \in \mathcal{L}, \bar{v}(A) \leq P(A) \leq v(A)\}$ where $\mathcal{L}$ is the set of probability measures on $S$ is known to be non-empty since $\bar{v}$ is convex (Shapley, 1971). This result will be used in Section 3.

**Example 1 continued** *Coming back to Example 1, let $\mathcal{P}$ be the set of all probability distributions on $S = \{1, 2, 3\}$ such that $P(\{1\}) = 1/3$, the set function defined by $v(A) = \sup_{P \in \mathcal{P}} P(A)$ for all $A \subseteq \{1, 2, 3\}$ and its dual $\bar{v}$ are given by:*

| $A$ | $\emptyset$ | $\{1\}$ | $\{2\}$ | $\{3\}$ | $\{1,2\}$ | $\{1,3\}$ | $\{2,3\}$ | $S$ |
|---|---|---|---|---|---|---|---|---|
| $v(A)$ | 0 | 1/3 | 2/3 | 2/3 | 1 | 1 | 2/3 | 1 |
| $\bar{v}(A)$ | 0 | 1/3 | 0 | 0 | 1/3 | 1/3 | 2/3 | 1 |

*In this case, it is easy to check that $v$ (resp. $\bar{v}$) is a concave (resp. convex) capacity. Moreover we have $core\{\bar{v}\} = \mathcal{P}$.*

The Choquet integral of a vector $x \in \mathbb{R}_+^m$ with respect to capacity $v$ is defined by:

$$C_v(x) = \sum_{i=1}^{m} [v(X_{(i)}) - v(X_{(i+1)})] x_{(i)} \quad (1)$$

$$= \sum_{i=1}^{m} [x_{(i)} - x_{(i-1)}] v(X_{(i)}) \quad (2)$$

where (.) represents a permutation on $\{1, \ldots, m\}$ such that $0 = x_{(0)} \leq x_{(1)} \leq \ldots \leq x_{(m)}$, $X_{(i)} = \{j \in S, x_j \geq x_{(i)}\} = \{(i),(i+1), \ldots, (m)\}$ for $i \leq m$ and $X_{(m+1)} = \emptyset$. Note that $X_{(i+1)} \subset X_{(i)}$, hence $v(X_{(i)}) \geq v(X_{(i+1)})$ for all $i$. The Choquet integral generalizes the classical notion of expectation with the following interpretation based on Equation (2): for a given vector $x = (x_1, \ldots, x_m)$, the outcome is at least $x_{(1)}$ with plausibility $v(X_{(1)}) = 1$, and the outcome might increase from $x_{(1)}$ to $x_{(2)}$ with plausibility $v(X_{(2)})$; the same applies from $x_{(2)}$ to $x_{(3)}$ with plausibility $v(X_{(3)})$, and so on. The overall integral is therefore obtained by aggregation of marginal increments $x_{(i)} - x_{(i-1)}$ weighted by plausibilities $v(X_{(i)})$.

In decision theory, the Choquet integral is often used in maximization problems under the form $C_v(u(x_1), \ldots, u(x_m))$ with utility function $u$ on payoffs to be maximized (CEU criterion, Schmeidler (1989)). In path-planning problems where costs replace payoffs, we need to reformulate the criterion using a *disutility* function to be minimized. Let $w : \mathbb{R}_+ \to [0, 1]$ be an increasing disutility function on costs such that $w(0) = 0$ and $w(M) = 1$. We introduce the *Choquet Expected Disutility* (CED), a function to be minimized over all feasible vectors $x \in \mathbb{R}_+^m$, defined by:

$$\psi_v^w(x) = C_v(w(x_1), \ldots, w(x_m)) \quad (3)$$

Note that CED includes classical expected disutility as a particular case. Indeed, whenever $v$ is additively decomposable, we have $v(A) = \sum_{i \in A} v_i$ for all $A \subseteq S$, where $v_i = v(\{i\})$. Hence $v(X_{(i)}) - v(X_{(i+1)}) = v_{(i)}$ for all $i$ and $\psi_v^w(x) = \sum_{i=1}^{m} v_{(i)} w(x_{(i)}) = \sum_{i=1}^{m} v_i w(x_i)$. When used with a non-additive capacity, it offers additional descriptive possibilities. As an illustration, let us continue Example 1.

**Example 1 continued** Assume that $w(0) = 0$, $w(100) = 1$. We get: $\psi_v^w((0, 100, 100)) = v(\{2,3\})$, $\psi_v^w(((100, 0, 100)) = v(\{1,3\})$, $\psi_v^w((0, 100, 0)) = v(\{2\})$, $\psi_v^w((100, 0, 0)) = v(\{1\})$. If $v$ is the concave capacity introduced above, we get $v(\{2,3\}) = 2/3 < 1 = v(\{1,3\})$ and $v(2) = 2/3 > 1/3 = v(1)$. Hence we get the desired preferences, i.e. $(0, 100, 100) \succ (100, 0, 100)$ and $(0, 100, 0) \prec (100, 0, 0)$. This example shows that $\psi_v^w$ is compatible with some uncertainty aversion.

## 2.3 Uncertainty aversion in CEU theory

Uncertainty aversion means intuitively that smoothing or averaging a cost vector makes the decision maker better off. A useful formalization of this idea is introduced in Chateauneuf and Tallon (2002) through an axiom named *"preference for diversification"* due to its interpretation in the context of portofolio man-agement. This axiom can be reformulate in our framework:

**Definition 1** *A preference $\succsim$ reveals uncertainty aversion if, for any $x^1, \ldots, x^n \in \mathbb{R}_+^m$, and $\alpha_1, \ldots, \alpha_n \geq 0$ such that for all $\sum_{i=1}^n \alpha_i = 1$, we have:*
$[x^1 \sim x^2 \sim \ldots \sim x^n] \Rightarrow \sum_{i=1}^n \alpha_i x^i \succsim x^k, \ k = 1, \ldots, n$

Interestingly enough, it is shown in Chateauneuf and Tallon (2002) that, within CEU theory, the above axiom on preference is equivalent to choosing a concave utility $u$ and a convex capacity $v$. The direct counterpart of this result in our context (minimization of Choquet expected disutility) says that we should use a *convex* disutility $w$ and a *concave* capacity $v$ to exhibit uncertainty aversion with $\psi_v^w$ as defined in Equation (3). For this reason, throughout the paper, we will assume that $w$ is convex and $v$ is concave.

**Example 2** *In a two-scenario instance with scenarios having approximatively the same plausibility, consider two solution paths with cost vectors $x = (10, 0)$ and $y = (0, 10)$ respectively. A decision maker averse to uncertainty might be indifferent between $x$ and $y$, i.e. $x \sim y$, but would prefer $z = (5, 5)$ to $x$ and $y$, i.e. $z \succ x$ and $z \succ y$. Such preferences are fully consistent with the idea of robustness. Suppose now that we have a convex disutility $w$ defined by $\forall t \in \mathbb{R}_+, w(t) = t^2/100$ and a concave capacity $v$ such that $v(\{1\}) = v(\{2\}) = 2/3$. Then $\psi_v^w(x) = \psi_v^w(y) = 2/3 > \psi_v^w(z) = 1/4$ which induces the desired preferences. Note that using a concave disutility $w$ such as $w(t) = \sqrt{t/10}$ gives $\psi_v^w(x) = \psi_v^w(y) = 0.67 < 0.71 = \psi_v^w(z)$, which is not compatible with uncertainty aversion. The same observation can be made with a convex capacity.*

We are now in position to formulate the central problem of this paper:

**The $\psi_v^w$-OPT problem.**
*Instance:* a state space graph $G = (N, A)$ as introduced in Subsection 2.1, an initial node $s$ and a set $\Gamma \subseteq N$ of goal nodes, a finite set of states $S$, a scenario-dependent valuation $c : A \times S \to \mathbb{R}_+$, and a concave capacity $v$ and a convex disutility $w$.
*Goal:* determine a $\psi_v^w$-optimal path among all paths in $\mathcal{P}(s, \Gamma)$ where $\psi_v^w$ is defined as in Equation (3).

### 2.4 Complexity issues

If $v(A) = 1$ for all non-empty $A \subseteq S$, then $\psi_v^w(x) = \sum_{i=1}^m \left[w(x_{(i)}) - w(x_{(i-1)})\right] v(X_{(i)}) = w(x_{(m)}) = \max_{i \in S} w(x_i)$. Hence $\psi_v^w$-OPT reduces to the min-max search problem, which was proved NP-hard by Murthy and Her (1992). This shows that $\psi_v^w$-OPT is also NP-hard.

We wish to propose a heuristic search algorithms to solve the $\psi_v^w$-OPT problem. The efficient resolution of shortest-path problems with $A^*$ in the classical case relies on the Bellman principle that justifies local pruning of sub-optimal subpaths during the search. However, in the case of multiple scenarios the CED model breaks the Bellman principle, as shown by the following example:

**Example 3** *Consider an instance with two paths $P$ and $P'$ from $s$ to a node $n$ with costs vectors $x = (0, 100, 0)$ and $x' = (100, 0, 0)$. Consider a Choquet expected disutility criterion $\psi_v^w$ with $w(0) = 0, w(100) = 1$ and $v(\{1\}) = 0.4, v(\{2\}) = 0.5, v(\{2, 3\}) = 0.7, v(\{1, 3\}) = 0.8$. Here we have $\psi_v^w(x) = 0.5$ and $\psi_v^w(x') = 0.4$ and therefore $P' \succ P$. For search efficiency, we might want to prune $P$ at node $n$ due to the existence of the better subpath $P'$. However, it might be a mistake. Assume, for example, that a path $P''$ from $n$ to $\Gamma$ exists with cost $y = (0, 0, 100)$; we have $P' \cup P'' \prec P \cup P''$ because $\psi_v^w(x' + y) = \psi_v^w((100, 0, 100)) = 0.8 > 0.7 = \psi_v^w((0, 100, 100)) = \psi_v^w(x + y)$.*

Such a preference reversal shows that, at any node visited during the search, a naive pruning of sub-optimal sub-paths with respect to $\psi_v^w$ might lose the admissibility of the algorithm. A similar violation of the Bellman principle is highlighted in De Cooman and Troffaes (2005), in the general framework of dynamic programming with uncertain gain and imprecise probabilities. It concerns the notion of $\underline{P}$-maximinity, a counterpart of $\psi_v^w$-optimality for imprecise probabilities. The next section presents two admissible algorithms to solve $\psi_v^w$-OPT.

## 3 ALGORITHMS

### 3.1 Optimistic Heuristics for $\psi_v^w - OPT$

Before introducing algorithms for the $\psi_v^w$-OPT problem, we establish an inequality providing an easily computable lower bound for criterion $\psi_v^w(x)$.

**Proposition 1** *If $v$ is concave then for all $P \in core(\bar{v})$ the following inequality holds: $\psi_v^w(x) \geq \sum_{i=1}^m p_i w(x_i)$ with $p_i = P(\{i\})$. Moreover, if $w$ is convex we have: $\psi_v^w(x) \geq w(\sum_{i=1}^m p_i x_i)$.*

**Proof.** $\psi_v^w(x) = \sum_{i=1}^m \left[w(x_{(i)}) - w(x_{(i-1)})\right] v(X_{(i)})$
$\geq \sum_{i=1}^m \left[w(x_{(i)}) - w(x_{(i-1)})\right] P(X_{(i)})$ since inequalities $v(X_{(i)}) \geq P(X_{(i)})$ hold for all $i$ ($P \in core(\bar{v})$).
Further, $\sum_{i=1}^m \left[w(x_{(i)}) - w(x_{(i-1)})\right] P(X_{(i)}) = \sum_{i=1}^m [P(X_{(i)}) - P(X_{(i+1)})] w(x_{(i)}) = \sum_{i=1}^m p_{(i)} w(x_{(i)})$
$= \sum_{i=1}^m p_i w(x_i) \geq w(\sum_{i=1}^m p_i x_i)$ when $w$ is convex. $\square$

Hence, any probability distribution $P$ in $core(\bar{v})$ can be used to produce a lower bound that can be used as an optimistic heuristic in the search. Among the natural choices for $P$ that give efficient bounds in practice, let us mention Shapley values $\phi_i$ that represent the average marginal contribution of any state $i = 1, \ldots, m$ to events (Shapley, 1971). Shapley values are positive and add up to one. In our context they are defined by:

$$\phi_i = \sum_{K \subseteq S \setminus \{i\}} \frac{(m - |K| - 1)! |K|!}{m!} (\bar{v}(K \cup \{i\}) - \bar{v}(K))$$

Another possible choice among probability laws in $core(\bar{v})$ is $P^*$, the probability distribution maximizing entropy $h(P) = -\sum_{i=1}^{m} p_i \log p_i$ over the $core(\bar{v})$. This propability distribution can easily be obtained using the following greedy algorithm proposed in Jaffray (1995), where $p_i^*$ stands for the probability $P^*(\{i\})$.

---
**Algorithm 1**: computing $P^*$ with max-entropy
---
Initialization:
$A \leftarrow \emptyset$;
$B \leftarrow \emptyset$;
**while** $B \neq S$ **do**
    $A \leftarrow \arg\min\{\frac{v(B \cup E) - v(B)}{|E|}, E \subseteq S \setminus B, E \neq \emptyset\}$;
    **for** all $i \in A$ **do**
        $p_i^* \leftarrow \frac{v(B \cup A) - v(B)}{|A|}$;
    $B \leftarrow B \cup A$;
**Output**: $(p_1^*, \ldots, p_m^*)$
---

**Example 1 continued** *Coming back to Example 1, Algorithm 1 outputs probabilities $p_1^* = p_2^* = p_3^* = 1/3$. With $w(0) = 0$ and $w(100) = 1$ we get the following evaluations for paths.*

| $x^k$ | 1 | 2 | 3 | $\psi_v^w$ | $\sum_{i=1}^{3} p_i^* x_i^k$ |
|---|---|---|---|---|---|
| $x^1$ | 0 | 100 | 100 | 2/3 | 2/3 |
| $x^2$ | 100 | 0 | 100 | 1 | 2/3 |
| $x^3$ | 0 | 100 | 0 | 2/3 | 1/3 |
| $x^4$ | 100 | 0 | 0 | 1/3 | 1/3 |

*We observe that $\psi_v^w(x^k) \geq \sum_{i=1}^{3} p_i^* x_i^k$ for all $k$.*

In Example 1, Shapley values coincide with maximal entropy probabilities but this is not true in the general case.

We introduce now two exact algorithms for the $\psi_v^w$-OPT problem. The first one is based on multiobjective search and the second one is based on a ranking approach with prior scalarization of the problem.

### 3.2 Multiobjective Search

The presence of $m$ scenarios in our problem provides $m$ different viewpoints on costs in the state space graph. This clearly points out connections with the field of multiobjective optimization. More precisely, since $\psi_v^w(x_1, \ldots, x_m)$ is increasing in each component, it is clear that $\psi_v^w$-optimal vectors belong to the Pareto set, i.e., the set of feasible vectors that cannot be improved on a component without being degraded on another. More formally, within a set $X \subseteq \mathbb{R}_+^m$, the Pareto-set is defined by:

$ND(X) = \{x \in X : \forall y \in X, \ y \succsim_P x \Rightarrow x \succsim_P y\}$

where $\succsim_P$ is the weak Pareto-dominance relation defined on $\mathbb{R}_+^m$ by: $x \succsim_P y \iff [\forall i \in S, x_i \leq y_i]$. Hence the search for a $\psi_v^w$-optimal path might be performed with a multiobjective search algorithm. For this reason, we recall now the main features of the $MOA^*$ algorithm proposed by Stewart and White III (1991) to find the set $ND(\mathcal{P}(s, \Gamma))$ of non-dominated solution paths in $G$. Algorithm $MOA^*$ relies on the same foundations as $A^*$, adapted to the field of vector-valued evaluation. In particular, the Bellman principle holds. Thus, any subpath from $s$ to a node $n \in N$ of a non-dominated solution path is non-dominated in $\mathcal{P}(s, n)$. Hence, the algorithm constructs incrementally the non-dominated solution paths from non-dominated subpaths.

In a graph valued by cost vectors, there may exist several non-dominated paths to reach a given node $n$. In $MOA^*$ these paths are kept in a label attached to the concerned node and all these paths are expanded when the label is developed. The choice of a node $n$ to be developed at the current iteration is based on a vector valued evaluation function that uses a vector-valued heuristic $H(n)$ containing cost vectors underestimating the real costs of any path in $\mathcal{P}(n, \Gamma)$. This set is an optimistic approximation of $H^*(n) = ND(\mathcal{P}(n, \Gamma))$.

Recently, a variant of this algorithm was proposed in Mandow and de la Cruz (2005), managing and expanding labels attached to paths rather than nodes. This variant makes it possible to not expand some sub-paths that will lead to dominated paths whereas they would be expanded by $MOA^*$. To this end, each detected subpath $P$ from $s$ to $n$, is given a label $l = [n_l, g_l, f_l, P_l]$, where $n_l = n$ is the terminal node of $P$, $g_l$ is the cost vector associated to $P$, $f_l$ is the cost vector $g_l + h(n)$ where $h(n)$ is a vector-valued heuristic of $H(n)$, and $P_l$ is the sequence of nodes $\langle s, \ldots, n \rangle$ forming $P$. In the sequel, $\mathcal{L}(n)$ will denote the set of labels attached to detected paths in $ND(\mathcal{P}(s, n))$. Two sets of labels are mantained to avoid the expansion of a label already expanded: the set of OPEN labels $O$ contains those labels attached to already detected but not expanded subpaths, and the set of CLOSED labels $C$ contains those labels already detected and expanded. The expansion of a label $l$, at a given step of the search, proceeds as follows: (1)

move $l$ from $O$ to $C$; (2) insert in $O$ new labels of type $[n', g_l + c(n, n'), g_l + c(n, n') + h, \langle P_l, n' \rangle]$ for all $n' \in S(n_l)$ and $h \in H(n')$.

At any iteration, the algorithm selects a new non-dominated label in $O$. It stops when there is no remaining label to expand (i.e. $O = \emptyset$), which means there is no remaining Pareto-optimal solution paths to explore.

Starting from this general multiobjective search procedure, we use two pruning rules to focus the search on $\psi_v^w$-optimal paths.

**RULE 1:** at node $n$, we prune any label $l$ in $\mathcal{L}(n) \cap O$ such that there exists another label $l' \in \mathcal{L}(n)$ such that $g_{l'} \succsim_P g_l$. Indeed, path $P_l$ cannot lead to a non-dominated solution path since any extension of $P_l$ with a sub-path $P$ will be dominated by path $P_{l'} \cup P$. Therefore, $P_l$ cannot lead to a $\psi_v^w$-optimal solution path either.

In order to introduce the second pruning rule, we consider, for any probability vector $p = (p_1, \ldots, p_m) \in core(\bar{v})$, an arc valuation function $c_p : A \to \mathbb{R}_+$ derived from $c$ by setting $c_p(a) = \sum_{i \in S} p_i c(a, i)$ for all $a \in A$ (linear aggregation of costs). Hence Rule 2 can be expressed as follows.

**RULE 2:** at node $n$ we discard any label $l$ such that $\max\{\psi_v^w(f_l), w(c_p(P_l) + \bar{h}(n_l))\} \geq \psi_v^w(x^*)$, where $x^*$ is the cost vector of the current $\psi_v^w$-optimal solution path among all already detected solution paths and $\bar{h}(n)$ is an optimistic heuristic on the cost of the best path from $n$ to a goal node with respect to $c_p$.

Rule 2 enables the early elimination of uninteresting labels while keeping admissibility of the algorithm, provided optimisic heuristic are used. Indeed, consider any solution path $P = P_l \cup P'$ that extends $P_l$. Then the two following cases must be considered.
i) $\psi_v^w(f_l) \geq \psi_v^w(x^*)$. Path $P$ has a cost vector $g$ which is Pareto-dominated by $f_l$ provided $H$ is admissible, i.e., $\forall n \in N, \forall h^*(n) \in H^*(n), \exists h \in H(n)$ such that $h(n) \succsim_P h^*(n)$. Thus we have $\psi_v^w(g) \geq \psi_v^w(f_l)$ and therefore $\psi_v^w(g) \geq \psi_v^w(x^*)$ by transitivity.
ii) $w(c_p(P_l) + \bar{h}(n_l)) \geq \psi_v^w(x^*)$. By Proposition 1, we have $\psi_v^w(g) \geq w(c_p(P))$ where $g$ is the cost vector of $P$; moreover $w(c_p(P)) = w(c_p(P_l) + c_p(P')) \geq w(c_p(P_l) + \bar{h}(n_l))$ since $\bar{h}$ is optimistic. Hence $\psi_v^w(g) \geq w(c_p(P_l) + \bar{h}(n_l)) \geq \psi_v^w(x^*)$.
In both cases we have $\psi_v^w(g) \geq \psi_v^w(x^*)$ which proves that $P$ is suboptimal and justifies Rule 2.

Algorithm 2 formally presents the resulting search procedure. In this procedure, $sol$ denotes the current $\psi_v^w$-optimal solution path represented by the pair (cost vector, sequence of nodes), and $ND(\mathcal{L}) = \{l \in \mathcal{L} : \forall k \in \mathcal{L}, g_k \succsim_P g_l \Rightarrow g_l \succsim_P g_k\}$.

**Algorithm 2**: Choquet-optimization
Initialization:
$C \leftarrow \emptyset$;
$\mathcal{L}(s) \leftarrow \bigcup_{h \in H(s)} \{[s; (0, \ldots, 0); h; \langle s \rangle)]\}$;
$O \leftarrow \{l' \in \mathcal{L}(s)\}$;
$l \leftarrow \arg\min_{k \in \mathcal{L}(s)} \max\{\psi_v^w(f_k), w(\bar{h}(s))\}$ ;
$\lambda \leftarrow \infty$;
**while** $[O \neq \emptyset, \psi_v^w(f_l) < \lambda, w(c_p(P_l) + \bar{h}(n_l)) < \lambda]$ **do**
    move $l$ from $O$ to $C$;
    **if** $n_l \in \Gamma$ **then**
        **if** $\psi_v^w(g_l) < \lambda$ **then**
            $sol \leftarrow (g_l, P_l)$;
            $\lambda \leftarrow \psi_v^w(g_l)$;
    **else**
        **for** each node $n' \in S(n_l)$ **do**
            **for** each cost vector $h(n') \in H(n')$ **do**
                $x \leftarrow g_l + c(n_l, n') + h(n')$;
                **if** $\psi_v^w(x)) < \lambda$ **then**
                    create label
                    $l' = [n', g_l + c(n_l, n'), x, \langle P_l, n' \rangle]$;
                    $\mathcal{L}(n') \leftarrow ND(\mathcal{L}(n') \cup \{l'\})$;
                    $O \leftarrow O \bigcup \{l'\}$
    $l \leftarrow \text{argmin}_{l' \in O} \max\{\psi_v^w(f_{l'}), w(c_p(P_{l'}) + \bar{h}(n_{l'}))\}$ ;
**Output**: $sol$

### 3.3 The Ranking Approach

For any path $P^j$ in the graph we have $c_p(P^j) = \sum_{a \in P^j} c_p(a) = \sum_{a \in P^j} \sum_{i \in S} p_i c(a, i) = \sum_{i \in S} p_i \sum_{a \in P^j} c(a, i) = \sum_{i \in S} p_i c(P^j, i)$. Let $x^j$ be the cost vector of $P^j$ defined by $x_i^j = c(P^j, i)$ for all $i$. Then the Choquet expected disutility associated to $P^j$ is $\psi_v^w(x^j) \geq w(\sum_{i \in S} p_i c(P^j, i))$ by Proposition 1, which is equal to $w(c_p(P^j))$. Therefore we have:

$$\psi_v^w(x^j) \geq w(c_p(P^j)) \qquad (4)$$

Hence the value of any path in the graph endowed with the scalar valuation $c_p$ provides a lower bound on the optimal value of $\psi_v^w(x)$ over feasible vectors $x$. This statement led us to seek the optimal solution of $\psi_v^w$-OPT through a ranking algorithm performing the enumeration of $k$ best paths by increasing values $c_p(P)$, i.e. by increasing values of $w(c_p(P))$. Denoting $P^1, \ldots, P^j$ the first solution paths in the enumeration, we have $c_p(P^1) \leq \ldots \leq c_p(P^j)$. Suppose that $P^j$ is the first of these paths satisfying $w(c_p(P^j)) \geq \psi_v^w(x^*)$ where $x^* = \arg\min_{i=1,\ldots,j} \psi_v^w(x^i)$ is the cost of the current $\psi_v^w$-optimal path $P^*$, then Equation (4) implies that all forthcoming paths $P^k, k > j$ in the enumeration will satisfy $\psi_v^w(x^k) \geq w(c_p(P^k))$. Since by construction $w(c_p(P^k)) \geq w(c_p(P^j)) \geq \psi_v^w(x^*)$, we get $\psi_v^w(x^k) \geq \psi_v^w(x^*)$. Therefore $P^*$ is a $\psi_v^w$-optimal solution path and we can stop the enumeration.

Let us explain now how to rank solution paths from best to worst according to $c_p$. We propose a modified version of A* that uses labels attached to paths. To any detected path $P$ with a cost vector $x$ from $s$ to $n \in N$, we assign a label $l = [n_l, g_l, \bar{g}_l, P_l]$ where $n_l$ is node $n$, $g_l$ is cost vector $x$, $\bar{g}_l$ is scalar cost $c_p(P)$, and $P_l$ is the sequence of nodes $\langle s, \ldots, n \rangle$ forming $P$. After the determination of a first $c_p$-optimal solution path, we have to continue so as to find the second-best solution path and then the following. For that, all detected sub-paths during the search (i.e., all labels) are kept. Therefore, several labels can be attached to a same node. Moreover, in order to favour a depth-first search in the state space graph, only the current most promising label on each node is expanded, until the corresponding solution path is found. To implement this principle, we use the two sets of nodes $O$ (set of open nodes), and $C$ (set of closed nodes). More precisely, set $O$ is the set of already detected nodes having no expanded label. Set $C$ is defined as the set of nodes having exactly one expanded label corresponding to a path which is not part of an already detected solution path.

The complete procedure is formalized in Algorithm 3 given below, where $\mathcal{L}(n)$ is the set of labels attached to $n$, $sol$ is the current $\psi_v^w$-optimal solution path represented by the ordered-pair (cost vector, sequence of nodes), and $\bar{h} : N \to \mathbb{R}$ is an optimistic heuristic on the cost of the best path from $n$ to a goal node with respect to $c_p$.

## 4 NUMERICAL EXPERIMENTS

We have performed numerical experiments of Algorithms 2 and 3 on path-planning problems, the size of which varies from 1,000 nodes to 3,000 (with arc density of 45%). On each transition, the cost of each scenario is randomly drawn between 0 and 100. For the two algorithms, the tests have been performed with the convex disutility function $w(x) = x^2$, and a concave capacity $v_1$ defined, for all $A \subseteq S$, as follows: $v_1(A) = 1 - (\sum_{i \notin A} p_i)^2$ for a randomly drawn probability distribution $(p_1, \ldots, p_m)$.

In Algorithm 2, heuristic $H(n)$ used at node $n$ is the vector $h(n)$ defined by $h_i(n) = \gamma h_i^*(n)$ where $h_i^*(n)$ is the cost of the shortest path in $\mathcal{P}(n, \Gamma)$ with respect to scenario $i \in S$ and $\gamma$ is randomly drawn in $[0.7; 1)$. We have tested Algorithm 3 with a heuristic function $\bar{h}$ at node $n$, determined by setting $\bar{h}(n) = \gamma \bar{h}^*(n)$ where $\bar{h}^*(n)$ is the cost of the shortest path in $\mathcal{P}(n, \Gamma)$ with respect to $c_p$, and $\gamma$ is randomly drawn in $[0.7; 1)$.

We compare here execution times of Algorithms 2 ($A2$) and 3 ($A3$) for two different lower-bounds obtained for the two probability vectors considered in Section 3.2 (maximal entropy ($p_i^*$) and Shapley values ($\phi_i$)). The results are given in Table 1. They show the efficiency of the two algorithms. Remark that execution times are slightly better with Algorithm 3 but the difference is not very significant. Moreover, probabilities $p_i^*$ used for linear scalarization seem to provide a slightly better bound than $\phi_i$ in most cases. We have performed the same experiments with another capacity $v_2$ defined, for all $A \subseteq S$, as follows: $v_2(A) = \sum_{E \cap A \neq \emptyset} \varphi(E)$ where $\{\varphi(E), E \subseteq S\}$ are randomly drawn positive Möbius masses adding up to 1, which ensures that $v_2$ is a plausibility function (for more details, see Shafer (1976);

**Algorithm 3**: Choquet-optimization by ranking
Initialization:
$C \leftarrow \emptyset$;
$O \leftarrow \{s\}$;
$\mathcal{L}(n) \leftarrow \emptyset, \forall n \in N$;
$l \leftarrow [s, (0, \ldots, 0), 0, \langle s \rangle]$;
$\mathcal{L}(s) \leftarrow \{l\}$;
$\lambda \leftarrow \infty$;
**while** $O \neq \emptyset$ and $w(\bar{g}_l + \bar{h}(n_l)) < \lambda$ **do**
    remove $l$ from $\mathcal{L}(n_l)$;
    remove $n_l$ from $O$ and put $n_l$ in $C$;
    **if** $n_l \in \Gamma$ **then**
        **if** $\psi_v^w(g_l) < \lambda$ **then**
            $sol \leftarrow (g_l, P_l)$;
            $\lambda \leftarrow \psi_v^w(g_l)$;
        **for** each node $n$ in $P_l$ **do**
            remove $n$ from $C$;
            **if** $\mathcal{L}(n) \neq \emptyset$ **then**
                put $n$ in $O$;
    **else**
        **for** each node $n \in S(n_l)$ **do**
            $l' \leftarrow [n, g_l + c(n_l, n), \bar{g}_l + \bar{c}(n_l, n), \langle P_l, n \rangle]$;
            **if** $\bar{g}'_l + \bar{h}(n) < \lambda$ **then**
                $\mathcal{L}(n) \leftarrow \mathcal{L}(n) \cup \{l'\}$;
                **if** $n \notin C$ **then**
                    put $n$ in $O$;
    $l \leftarrow \mathrm{argmin}\,\{\bar{g}'_l + \bar{h}(n_{l'}), l' \in \bigcup_{n \in O} \mathcal{L}(n)\}$;
**Output**: $sol$

Table 1: Execution times (s) of Algorithms

|   |  |N| | 3 scen. $p_i^*$ | 3 scen. $\phi_i$ | 5 scen. $p_i^*$ | 5 scen. $\phi_i$ | 10 scen. $p_i^*$ | 10 scen. $\phi_i$ |
|---|---|---|---|---|---|---|---|
| $A2$ | 1,000 | 0 | 0.1 | 0.1 | 0.1 | 0.6 | 0.9 |
|  | 2,000 | 0.1 | 0.2 | 0.2 | 0.5 | 1.7 | 3.3 |
|  | 3,000 | 0.2 | 0.3 | 0.4 | 0.8 | 3.5 | 6.2 |
| $A3$ | 1,000 | 0 | 0.1 | 0 | 0.1 | 0.2 | 0.3 |
|  | 2,000 | 0.1 | 0.3 | 0.1 | 0.5 | 0.6 | 1.1 |
|  | 3,000 | 0.1 | 0.4 | 0.2 | 0.7 | 1.3 | 3.6 |

Chateauneuf and Jaffray (1995)). The execution times were similar.

## 5 FUTURE WORK

We have shown the potential of the Choquet integral in modelling risk-averse preferences for path-planning under uncertainty. Despite the complexity of the $\psi_v^w$-OPT problem in the worst case, the experiments show that the two heuristic search algorithms introduced in the paper are very efficient on average. In the future, it might be interesting to investigate the use of the Choquet integral in dynamic decision making problems e.g. decision trees or Markov decision processes. The main problem there is the existence of dynamic inconsistencies induced by the Choquet integral in sequential decision making.

Another direction might be to explore the potential of the Choquet integral in multiobjective planning problems. As shown by Grabisch (1996), the descriptive potential of the Choquet integral provides interesting possibilities in the field of multicriteria analysis. We might use the Choquet integral to characterize fine compromise solution paths in state space graphs. The algorithmic material presented here could certainly be adapted to determine Choquet-optimal paths in such multiobjective search problems.

### Acknowledgements

We would like to thank Jean-Yves Jaffray and Judy Goldsmith for fruitful discussions on this work and anonymous reviewers for their useful comments. This work has been supported by the ANR (project PHAC) which is gratefully acknowledged.